# Recognizing Micro-Expression in Video Clip with Adaptive Key-Frame Mining

Min Peng[1,2,†], Chongyang Wang[3,†*], Yuan Gao[4], Tao Bi[3], Tong Chen[5], Yu Shi[1], Xiang-Dong Zhou[1]

*Abstract*—As a spontaneous expression of emotion on face, micro-expression reveals the underlying emotion that cannot be controlled by human. In micro-expression, facial movement is transient and sparsely localized through time. However, the existing representation based on various deep learning techniques learned from a full video clip is usually redundant. In addition, methods utilizing the single apex frame of each video clip require expert annotations and sacrifice the temporal dynamics. To simultaneously localize and recognize such fleeting facial movements, we propose a novel end-to-end deep learning architecture, referred to as adaptive key-frame mining network (AKMNet). Operating on the video clip of micro-expression, AKMNet is able to learn discriminative spatio-temporal representation by combining spatial features of self-learned local key frames and their global-temporal dynamics. Theoretical analysis and empirical evaluation show that the proposed approach improved recognition accuracy in comparison with state-of-the-art methods on multiple benchmark datasets.

## I. INTRODUCTION

PREVISOULY, as an obvious facial movement pattern, macro-expression is one of the dominant modalities for non-contact emotion recognition. Macro-expression is consciously expressed and typically lasts from 1/2 to 4 seconds, which can be easily perceived by normal people [1]. In addition to understanding the macro-expression, many works have been conducted on identifying and recognizing them (for review, see [2] [3]). However, it has been latterly pointed out by several studies [4] [5] that macro-expression could be misleading and leveraged to conceal the genuine hidden emotion. In contrast, micro-expression is found to be expressed unconsciously as a transient (last less than 1/2 second) and subtle facial movement and proved to reveal the underlying emotions [5]. Due to the above reasons, the recognition of micro-expression has become a new popularity in the emotion recognition domain and boosted a wide range of real-world applications [4] [6] [7].

Research on automatic recognition of micro-expression first took the advantage of abundant works on macro-expression recognition, where the popular pipeline is, e.g., image preprocessing including normalization and cropping etc., and feature engineering followed up by classification with machine learning methods. To name a few, Zhao et al. [8] [16] adopted the local binary pattern (LBP) onto the three orthogonal planes (LBP-TOP) of a micro-expression clip, to extract the dynamic facial texture features and applied support vector machine (SVM) for the classification. Another adaption of LBP-TOP is seen in the work by Huang et al [34], where the spatio-temporal completed quantization patterns (STCLQP) was proposed to consider sign, magnitude, and orientation as feature components. Vector quantization was adopted to obtain compact and discriminative codebooks in spatiotemporal domain. To reduce the information redundancy existed in LBP-TOP features, Wang et al. [9] further designed an algorithm called LBP-six intersection points (LBP-SIP) to replace LBP-TOP for the feature extraction process. After this, Wang et al [33] proposed a tensor independent color space (TICS) to replace the traditional RGB color space. Their experiments found that perceptual color spaces like CIELab and CIELuv were better than RGB to represent the dynamic texture information of micro-expression. In addition, Li et al. [10] conducted a comparison among LBP-TOP, histogram of oriented gradients (HOG), and histogram of image gradient orientation (HIGO) about their feature extraction capability on micro-expression videos.

Except for these evolved from past researches on facial macro expression recognition, methods for generic video analysis from a wider computer-vision community have also inspired the research on facial micro-expression recognition. More recently, Optical Flow [11] has been harnessed to enable better temporal feature extraction of micro-expression. Happy et al [35] studied the temporal feature extraction of facial micro movements with fuzzy histogram of optical-flow orientation (FHOFO) features. Liu et al. [12] divided the face into 36 regions based on the facial action coding system [38], and captured the subtle facial movement in video clip by computing the main directional mean optical-flow (MDMO) feature of each area. In their following work, it is also tried to add sparsity into such feature extraction to reduce redundancy [36]. Later, Xu et al. [13], similarly based on optical flow, designed a fine-grained temporal alignment method and an optical-flow

Manuscript submitted for review in March 2021.

[1]Chongqing Institute of Green and Intelligent Technology, Chinese Academy of Sciences, Chongqing, China. (e-mail: {pengmin, shiyu, zhouxiangdong}@cigit.ac.cn).

[2]Chongqing School, University of Chinese Academy of Sciences, Chongqing, China.

[3]UCL interaction centre, University College London, London, United Kingdom. (e-mail: {chongyang.wang.17, t.bi}@ucl.ac.uk).

[4]Shenzhen Institute of Artificial Intelligence and Robotics for Society, Shenzhen, China. (email: gaoyuan@cuhk.edu.cn).

[5]School of Electronic and Information Engineering, Southwest University, Chongqing, China. (email: c_tong@swu.edu.cn).

[†]Equal contribution.

[*]Correspondence author.

direction optimization strategy for the feature extraction of micro-expression clips. So far, accuracies on micro-expression recognition achieved with these feature-engineering methods are slightly higher than the level of what the human expert can achieve. However, sophisticated pre-processing and feature computation stages are needed which are time-consuming.

One commonality of the studies mentioned above is the use of an entire video clip as input for the recognition. However, such practices have been challenged from several perspectives. The micro-expression clips provided in benchmark datasets, e.g., CASME I [14], CASME II [15], SMIC [42], and SAMM [46] are manually annotated, providing the location of onset, apex, and offset frames. It should be noted that we deem the original video clip (starts from onset frame and ends with offset frame) of each participant containing a single type of micro-expression provided in the dataset without extra segmentation information as the ***raw video clip***. The number of frames between the onset and offset frames of each video clip ranges from 9 to more than 100. As a result, temporal normalization technique like the temporal interpolation method (TIM) [8] has become a popular prior step for many existing methods, which was originally designed to cope with video clips of different lengths. However, such method would disturb the temporal order of the original video clip to some extent (e.g., making a temporal process longer or shorter), while a better model should accept input with varying temporal lengths. Additionally, empirical studies [19] [20] showed that most of the frames within a micro-expression clip are not contributing to the recognition, which can be referred to as redundant frames. Such result is also due to the fact that micro-expression itself is transient and mostly expressed within several local frames, which can be deemed as the ***key frames***. Moreover, two recent studies [21] [22] supported this argument by showing the advantage of using the apex frame alone for the recognition instead of the entire clip. Still, two notable disadvantages of apex frame-based methods are: i) the need for extra manual annotation of the apex frame, which reduces the end-to-end integrity of a system; ii) the loss of the temporal pattern of micro-expression, which could hinder the generalization capability of a system, e.g., on the recognition of an emotion type across different participants. A recent work by Peng et al. [27] proposed to improve the latter issue by integrating the temporal information adjacent to the apex frame. In [41], the author tried to enrich the feature from the apex frame by searching for useful extra information with genetic algorithms. However, annotations of the apex frame are still needed.

In this work, we propose an end-to-end deep learning architecture named **a**daptive **k**ey-frame **m**ining **net**work (AKMNet), which is able to adaptively learn the key frames and simultaneously extract the spatio-temporal features for micro-expression recognition given a raw video clip. Specifically, AKMNet has three modules, namely the traversal-processing, adaptive key-frame mining, and spatio-temporal fusing modules. The traversal-processing module is a spatial feature extraction unit, which processes all frames of the input video in a traversal manner, aimed to create a sequence of spatial features accordingly. Then, in key-frame mining module, a temporal subset comprising the key frames is learned by searching through such spatial feature sequence. This is achieved using a constrained optimization process with designed loss functions. Finally, in spatio-temporal fusing module, this subset of spatial features will be further processed to acquire the spatio-temporal feature matrix, on which the classification is performed. In the training of AKMNet, these three modules are integrated to conduct end-to-end learning for micro-expression recognition. The location of key frames that contribute most to the recognition may change per participants and micro-expression categories, and can be leveraged for additional micro-expression spotting.

To the best of our knowledge, this is the first work deliberated to integrate end-to-end learning of the informative temporal subset from a video clip and the recognition of micro-expression in a single network. Also, the design of all modules in the proposed network is independent on the length of input video clip. In other words, the network tolerates micro-expression clip of various lengths. The contributions of this paper are three-fold:

-A novel deep learning architecture, referred to as AKMNet, is proposed for micro-expression recognition, featuring the ability of mining the temporal subset of key frames that contribute most to the recognition performance.

-The theoretical space of the apex frame is enriched via comparing it with the key frames learned by the AKMNet in a pure data-driven manner.

-Through extensive experiments, the proposed AKMNet is shown to be effective for micro-expression recognition with noticeably improved performance, and it is generalizable to the recognition of macro-expression in video clips.

## II. RELATED WORKS

Following the feature-engineering methods mentioned in the last section, the past three years have witnessed a growing response in micro-expression researches towards the boom of deep learning in computer vision community. Various deep neural networks have been used in several studies. Without hand-crafted feature engineering, an end-to-end neural network model is capable of classification or prediction by learning from large sets of high dimensional (and low-level representation of) data [23]. For the research on micro-expression, efforts have been made to adapt several successful CNN and RNN models from the wider computer version domain to solve problems in this scenario.

First, researchers have focused on extracting representation from the entire micro-expression clip (including but not limited to frames between the onset and offset frames). To name a few, Peng et al. [18] proposed to use 3D-CNN to learn representation from the optical-flow sequence transformed from such clip, while SVM was used for the final classification. The proposed architecture possesses two streams, which were aimed to consume the data in different original framerates from the two benchmark datasets (CASME I/II), and each stream was also mitigated to only have 4 convolutional layers each followed by a pooling layer in order to alleviate the over-fitting risk. Through end-to-end implementation and testing on

CASMEI/II, such method achieved a best f1 score of 0.6667 with 3-fold cross validation, which is around 10% higher than previously state-of-the-art methods ([12] [13]). Considering the temporal dynamic of the video data, Khor et al [24] proposed an architecture with convolutional and recurrent layers. Together with the full video clip, the optical flow feature was also extracted at each timestep for a given period to provide richer temporal information. However, the accuracy they achieved was somehow limited, which may be due to the use of a complex network on the small datasets. In the two studies mentioned above, two popular network architectures (3D-CNN and CNN-RNN) originally used for video classification have been tested on micro-expression datasets, while the accuracy achieved is comparable if not better than previous hand-crafted feature engineering methods, under a practical end-to-end design with higher real-life impact.

However, except for the limited size of micro-expression datasets for deep learning models, another challenge existed here is also the redundancy introduced by unimportant frames within each video clip of micro-expression. Toward this issue, recent studies [21] [22] have explored the feasibility of using the apex frame of each video clip alone for the recognition, where the intensity of a micro-expression is believed to reach the highest. Both of the works proved that not all the frames of a micro-expression clip contribute to the recognition. Following up such finding, researchers begin to use the apex frame to conduct micro-expression recognition. Peng et al [19] proposed to use transfer learning to aid the micro-expression recognition on apex frames by pre-training on macro-expression data. As suggested in this work, micro-expression recognition on the apex frame opened a way to transfer knowledge learned from data of macro-expression, since both tasks are strongly relevant and operated on static images. Another work by Wang et al [20] proposed to use attention mechanism [25] to help the network focus on most informative spatial areas of an apex frame, as micro-expression is also spatially localized on small areas of face. Later on, another study by Peng et al [27] found that integrating the spatial information of an apex frame with temporal information of the frames adjacent to it produced higher generalization performance across different datasets. Furthermore, Xia et al [47] computed and utilized the optical flow feature between the onset and apex frames for micro-expression recognition using a recurrent convolutional network. Similarly, Liong et al [50] applied a three-stream network on such feature to improve the extraction of spatio-temporal features. One obvious limitation we find of the apex-frame based studies is the lack of a generic method to locate the essential frames e.g., the apex and onset frames. Some studies [21] [22] located apex frames based on spatial or spectral feature engineering, while the rest [19] [20] [27] relied on manual annotations. On the other hand, the way the temporal information is processed has been limited to, e.g., traversal processing of all the input frames, while a study on heuristic temporal information searching and extraction is still missing.

In this paper, to automatically deal with the redundancy of using all the frames in the micro-expression clip and the dependency on manual annotations of the apex frame, we propose to automatically learn the informative subset of key frames that contribute most to the recognition performance. Our method combines the automatic subset-searching, spatio-temporal representation fusion, and recognition, by proposing an end-to-end network called AKMNet. In the next section, the methodology is described in detail to show how we achieve this.

## III. METHODOLOGY

In this section, we first illustrate the overall architecture of the proposed AKMNet. We show how the proposed network is designed to automatically learn the key frames from a video clip of micro-expression. Then we describe in detail the proposed computational algorithms used in the adaptive key-frame mining module within AKMNet.

### A. Overview of AKMNet

Micro-expression is a transient and temporally localized facial movement indicating the internal emotion of a person. Therefore, the number of frames informative to the recognition of micro-expression is limited. Additionally, the duration of a micro-expression clip varies across different subjects, even for the same emotion. Such temporal nature and individual differences manifest that it is essential to first find the most informative frames, then to conduct the recognition thereon. Therefore, first, for a video clip of micro-expression, richer information are stored in some key frames in comparison with other redundant frames. Second, the distribution of the key frames could be sparse, thus each key frame could be connected by several redundant frames.

An overview of the proposed AKMNet is shown in Figure 1. Given a video input $V = \{I_1, I_2, ..., I_T\}$, with $I_T$ represents the $T$-th frame, the AKMNet first extracts the spatial feature for each frame using a ResNet-18 network [31]. The generated feature sequence is $F \in \mathbb{R}^{T \times C \times W \times H}$ where $F = \{P_1, P_2, ..., P_T\}$. $P_T \in \mathbb{R}^{C \times W \times H}$ is the spatial feature extracted from the $T$-th frame $I_T$, where $C, W, H$ are the number of channels, width, and height, respectively. The same ResNet-18 network is shared in the spatial-feature extraction across frames. Such is to i) ensure the feature extracted from each frame provides the same level of information; ii) tolerate clip input of different lengths; iii) reduce the number of trainable parameters so as to alleviate the overfitting risk on the small micro-expression datasets. The ResNet-18 network was first initialized using the ImageNet [37] and further pre-trained with a macro-expression dataset named AffectNet [39]. Such is aimed to transferring the ability of ResNet from generic image classification to facial expression recognition as well as to alleviate the over-fitting risk. Details for the adaptive key-frame mining module is given in the next subsection. Such module produces a feature sequence subset of $N$ frames that are expected to be most informative for the recognition, denoted as $F^{Key} \in \mathbb{R}^{N \times C \times W \times H}$, $F^{Key} = \{P_{k_1}, P_{k_2}, ..., P_{k_N}\}$. These key frames are listed following their original temporal order in the spatial feature sequence $F$, i.e. we have $1 \leq k_1 < k_N \leq T$. The number of key frames $N$ is adaptively changed given different input clips.

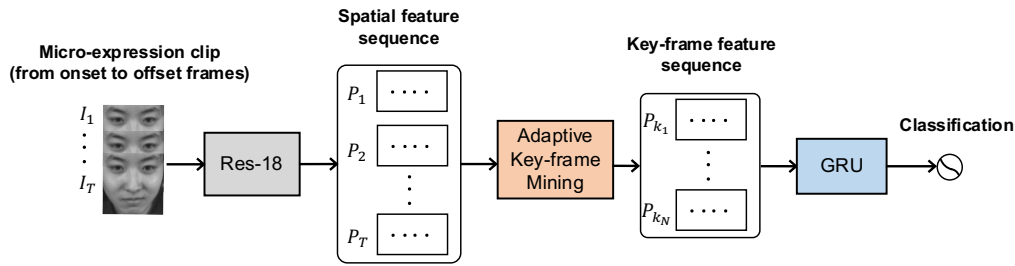

Fig. 1. Overview of proposed AKMNet. The network first conducts spatial feature extraction with ResNet-18 on the micro-expression clip input with frames number of T. Then N selected frames will be generated from the adaptive key-frame mining module which are further processed by bi-GRU networks to learn the spatio-temporal feature. The classification is conducted on the top of such feature.

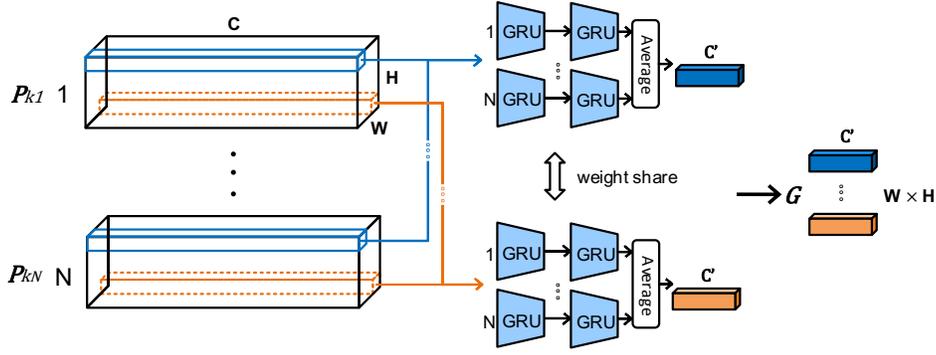

Fig. 2. The pixel-wise recurrent processing on key-frame feature sequence with bi-GRU networks. Each bi-GRU network processes a sequence of pixel-wise feature vectors extracted from all the key frames.

As shown in Figure 2, the key-frame feature sequence $F^{Key} = \{P_{k_1}, P_{k_2}, ..., P_{k_N}\}$ is processed by bi-directional GRU networks (bi-GRU) [32] to extract the temporal feature. Such bi-directional recurrent network is leveraged to extract the temporal context information, essential for micro-expression recognition. In addition, a GRU unit has fewer trainable parameters than a LSTM unit that is commonly used for temporal modeling, helpful to alleviate the overfitting risk on smaller datasets. Specially, the bi-GRU network is employed to conduct a **pixel-wise recurrent processing**. A single bi-GRU network is operating on a sequence of feature pixels with size of $N \times C$, extracted from the same position across all the key frames. As the size of a spatial feature $P_{k_N}$ is $C \times W \times H$, we have totally $W \times H$ bi-GRU networks in order to process all the feature pixel across all the frames. In this way, the temporal feature is extracted without pooling. Such is also aimed to maintain the essential spatial information of micro-expression. All the bi-GRU networks share the same set of learnable parameters, so as to help reduce parameter size and tolerate clip inputs of different lengths. For the bi-GRU network, given a sequential input at spatial position $(i, j)$ across all key frames $F^{Key}_{(i,j)} = \{P^{(i,j)}_{k_1}, P^{(i,j)}_{k_2}, ..., P^{(i,j)}_{k_N}\}$, the output $h_n$ at $n$-th key frame $P^{(i,j)}_{k_n}$ is computed as

$$h_n = (1 - z_n) * h_{n-1} + z_n * \tilde{h}_n, \quad (1)$$

$$z_n = \sigma\left(W_z \cdot [h_{n-1}, P^{(i,j)}_{k_n}]\right), \quad (2)$$

$$\tilde{h}_n = tanh\left(W_{\tilde{h}} \cdot \left[\sigma\left(W_r \cdot [h_{n-1}, P^{(i,j)}_{k_n}]\right) * h_{n-1}, P^{(i,j)}_{k_n}\right]\right), \quad (3)$$

where $[\cdot]$ represent the stacking of different vectors, $*$ is element-wise multiplication, $\sigma$ is sigmoid activation, $h_{n-1}$ is the output of the last key frame, $z_n$, and $\tilde{h}_n$ are the intermediate feature matrices, $W_z$, $W_{\tilde{h}}$, and $W_r$ are the trainable weight matrices of the bi-GRU network. The output of each bi-GRU network is the average of the output $h_n$ across all key frames. Such output from each bi-GRU network is concatenated to produce the **key-frame spatio-temporal representation** $G \in \mathbb{R}^{C' \times W \times H}$, where $C'$ is channels number of the feature vector produced by each bi-GRU network.

In our end-to-end training, a softmax layer is adopted thereon to conduct the final classification. The loss function used here can be written as

$$\mathcal{L}_c = -\sum_{m=1}^{M} \mathbf{1}\{y^{(G)} = m^*\} \log \frac{e^{W_{m'}G}}{\sum_{m=1}^{M} e^{W_m G}}, 1 < m \leq M, \quad (4)$$

where, $M$ is the number of the total classes, $m'$ indicates the groundtruth class of current input video clip, $y^{(G)}$ is the classified category identity of input feature $G$, $W$ is weight vector between output layer and previous layer. $\mathbf{1}\{\cdot\}$ is the eigenfunction, when $\{\cdot\}$ is true it will return 1, and 0 vice versa.

### B. Adaptive Key-Frame Mining Module

Here, we describe in detail how the key-frame mining module functions. In general, the key-frame mining module learns a function $f_{AKM}: F \to F^{key}$. A demonstration of the key-frame mining module is provided in Figure 3. As shown, the input to such module is the feature sequence $F \in \mathbb{R}^{T \times C \times W \times H}$, which is first processed with a global spatial pooling layer to produce a pooled feature sequence $\overline{F} \in \mathbb{R}^{T \times C}$. That is, we compress each feature map $P_t \in \mathbb{R}^{C \times W \times H}$ in the sequence $F$ into a $C \times 1$ feature vector $\overline{P}_t$, $t \in \{1, 2, ..., T\}$, which contains compact spatial information of each frame.

In order to acquire the key-frame sequence, a three-step key-frame mining module is proposed. First, a **local self-attention learning step** is adopted, where a coarse attentional weight $\alpha_t$ is assigned to each feature vector $\overline{P}_t$. Second, based on these attentional weights computed in the first step, a global feature vector $\tilde{P}$ is produced by aggregating all the feature vectors. Thereon, a fine-grained weight for each feature vector is acquired by computing the cosine correlation between each feature vector and the global feature vector. We call such as **global correlation learning step**. Finally, based on these correlation values, a sparse binary correlation vector $B$ is learned using a global sparsity and maximum mean-margin loss function. This is referred to as **sparse selection step**. These three steps proposed for the key-frame mining module are empirically justified in the next section to understand their individual impact.

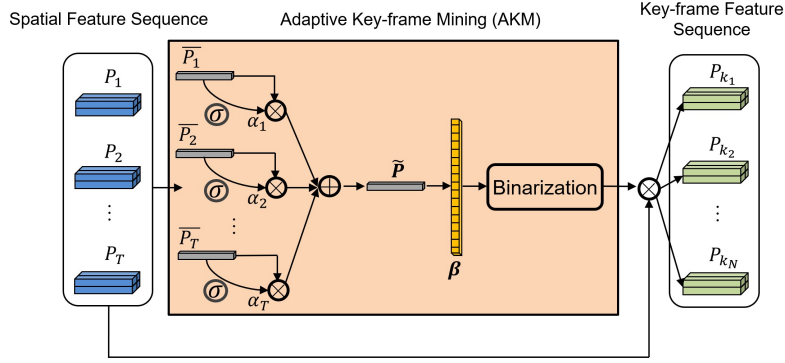

Fig. 3. The adaptive key-frame mining module.

In the local self-attention learning step, the local weight $\alpha_t$ for each frame feature input $\overline{P}_t$ is computed through a fully-connected layer with sigmoid activation function as

$$\alpha_t = \sigma(\boldsymbol{W}^T \overline{P}_t), \tag{5}$$

where $\sigma$ is sigmoid activation function, $\boldsymbol{W}^T$ is the trainable weight matrix for this layer.

In the global correlation learning step, we first acquire the global feature vector $\tilde{P}$ by aggregating all the input feature vectors and respective local self-attentional weights as

$$\tilde{P} = \sum_{t=1}^{T} \alpha_t \overline{P}_t. \tag{6}$$

Then, the fine-grained weight $\beta_t$ for each feature vector $\overline{P}_t$ is computed based on cosine correlation as

$$\beta_t = \frac{\overline{P}_t \cdot \tilde{P}}{\|\overline{P}_t\|_2 \|\tilde{P}\|_2}, \tag{7}$$

where $\|\cdot\|_2$ denotes the L2-norm.

Finally, in the sparse selection step, we implement the concept reasoned earlier that most frames in a video clip of micro-expression are redundant while the most informative frames are sparsely localized. Thus, we add sparsity to the weight vector $\{\beta_1, \beta_2, \ldots, \beta_t\}$. Inspired by the binarization process presented in the binary neural network [17] [26], we compute the binary index vector $B \in \{0,1\}^{T\times 1}$ as

$$B_t = \begin{cases} 1, & \beta_t > \frac{1}{T}\sum_{t=1}^{T}\beta_t \\ 0, & \beta_t \leq \frac{1}{T}\sum_{t=1}^{T}\beta_t \end{cases}, t \in \{1,2,\ldots,T\}, \tag{8}$$

where the location of each key frame is marked with a number 1, and 0 vice versa. With index vector $B$, the key frame feature sequence $F^{key}$ is acquired by

$$\boldsymbol{F}^{key} = \boldsymbol{F} \odot \boldsymbol{B}, \tag{9}$$

where $\odot$ denotes the tensor-fraction product and is conducted along the temporal dimension of the original feature sequence $\boldsymbol{F}$. To further enable the learning of the fine-grained weight $\beta_t$ and the sparsenization process, we propose to use a **global sparsity and maximum mean-margin (GS-MMM) loss function**. The first part of global sparsity loss is written as

$$\mathcal{L}_{GS} = max(0, sum(B) - 1), \tag{10}$$

where $max(\cdot)$ and $sum(\cdot)$ are the computation for maximum and sum, respectively. When $\mathcal{L}_{GS}$ is reduced, we make sure at least one key frame is selected. Another part of maximum mean-margin loss is written as

$$\mathcal{L}_{MMM} = C - (\frac{1}{sum(B)}\sum_{t=1|B_t=1}^{T}\beta_t - \frac{1}{T-sum(B)}\sum_{t=1|B_t=0}^{T}\beta_t), \tag{11}$$

where $C$ is set to 2 given that the value of fine-grained weight $\beta_t$ is within the range of cosine function. This loss function is aimed to further help discriminate the key frames from the others by assigning higher weights thereon, while avoid the trivial solution possibly led to by $\mathcal{L}_{GS}$. Thereon, the GS-MMM loss function is written as

$$\mathcal{L}_{GS-MMM} = \lambda_1 \mathcal{L}_{GS} + \lambda_2 \mathcal{L}_{MMM}, \tag{12}$$

where $\lambda_1$ and $\lambda_2$ are the weight constants controlling the impact of each loss, respectively. During the end-to-end training of the AKMNet, to combine the Equation 4 and 12, the final loss function $\boldsymbol{\Omega}$ to be reduced can be written as

$$\boldsymbol{\Omega} = \mathcal{L}_c + \mathcal{L}_{GS-MMM}. \tag{13}$$

In training, the network needs to learn the optimal key-frame subset that leads to the minimum of loss $\boldsymbol{\Omega}$, which shall also give the best classification performance. Through back propagation, the trainable weight matrices in the ResNet-18 and bi-GRU networks are updated. In the next subsection, we discuss how to transform such adaptive key-frame mining approach into a trainable block of the deep learning network.

### C. Weight Update Strategy

Normally, gradient descent is used to update a neural network, which requires all the computation conducted in the network to be differentiable and convex. However, the adaptive key-frame mining module described in the last subsection appears to be a non-convex process. Specifically, the calculation of binary index vector $B$ does not allow gradient computation. To make the process trainable, we develop an alternative minimization method to especially update the Equation 8 and 9.

We first define that, during the back-propagation, the gradient passed from the next layer to the layer of $F^{key}$ is denoted as $\frac{\partial \Omega}{\partial F^{key}}$. According to the chain rule, the gradient update for Equation 9 can be written as

$$\frac{\partial \Omega}{\partial F} = \frac{\partial \Omega}{\partial F^{key}} \cdot \frac{\partial F^{key}}{\partial F} = \frac{\partial \Omega}{\partial F^{key}} \cdot B. \quad (14)$$

vice versa

$$\frac{\partial \Omega}{\partial B} = \frac{\partial \Omega}{\partial F^{key}} \cdot \frac{\partial F^{key}}{\partial B} = \frac{\partial \Omega}{\partial F^{key}} \cdot F. \quad (15)$$

For Equation 8, the computation of index vector $B$ produces constant value, which does not allow gradient computation. Therefore, we modify such computation by replacing the constant value with the original fine-grained weight value, respectively. The Equation 8 can be re-written as

$$B_t = \begin{cases} \beta_t, & \beta_t > \frac{1}{T}\sum_{t=1}^{T}\beta_t \\ 0, & otherwise \end{cases}, t \in \{1,2,\dots,T\}. \quad (16)$$

Then, according to the chain rule, the gradient update for Equation 8 can be written as

$$\frac{\partial \Omega}{\partial \beta} = \begin{cases} \frac{\partial \Omega}{\partial B} \cdot \frac{\partial B}{\partial \beta} = \frac{\partial \Omega}{\partial B}, & \beta_t > \frac{1}{T}\sum_{t=1}^{T}\beta_t \\ 0, & otherwise \end{cases}, t \in \{1,2,\dots,T\}. \quad (17)$$

With Equation 14-17, the back-propagation for the weight updating of the proposed adaptive key-frame mining module is enabled.

*D. End-to-End Training of AKMNet*

In AKMNet, the recognition of micro-expression given a video clip input comprises three stages: i) the traversal spatial-feature extraction with ResNet-18 for the input video clip; ii) the adaptive key-frame mining with the proposed module, where a sequence of key feature maps is produced that are most valuable for the recognition; iii) learning the pixel-wise temporal features with the bi-GRU networks followed by feature concatenation, and classification. The end-to-end learning is enabled for the entire network. The training scheme is written in Algorithm 1[*].

## IV. EXPERIMENTS AND EVALUATIONS

In this section, we evaluate the proposed approach using several empirical experiments and theoretical analyses. We first describe the data pre-processing and experimental settings. Finally, we report and discuss the results.

*A. Data Preparation*

In our experiments, four benchmark datasets are used, namely CASME I [14], CASME II [15], SMIC [42], and SAMM [46]. Details for these datasets used in this work are summarized in Table I. Following previous works [12] [18] [47] [48], we merged original classes of CASME I/II and SAMM datasets into four categories, namely positive (happiness), negative (disgust, sadness, contempt, anger, and fear), surprise and others (tense, and repression). The original three categories of

**Algorithm 1** The End-to-End Training of AKMNet.
1. **Input:** $V = \{I_1, I_2, \dots, I_T\}$
2. **while** *training* **do**
3.   Extract the sequence of spatial feature $F = \{P_1, P_2, \dots, P_T\}$ from the input $V$, using the ResNet-18 network initialized with ImageNet and pre-trained with larger macro-expression datasets.
4.   With $F$, use the adaptive key-frame mining module to produce the index vector $B \in \{0, 1\}^{T \times 1}$ as Eq. 8.
5.   With $B$, generate the sequence of key-frame feature maps $F^{key} = \{P_{k_1}, P_{k_2}, \dots, P_{k_N}\}_{k_1 \geq 1}^{k_N \leq T}$ as Eq. 9.
6.   With $F^{key}$, first compute the temporal feature with a bi-GRU Network. Then produce a spatio-temporal feature matrix $G$ after feature concatenation.
7.   In the classification layer, compute loss $\Omega$ as Eq. 13.
8.   Based on $\Omega$, update the bi-GRU network, AKM module, and ResNet-18 network with back propagation.
9. **End**
10. **Output:** classification result $m$

SMIC dataset is unchanged. For all the methods compared in this work, the video clips comprising the frames between onset and offset frames provided in each dataset is used. Face detection and alignment is also conducted using the method adopted in [14] [15]. Additionally, a spatial augmentation method, referred to as *phase-based video magnification method* [49], is used for all the video clips. Such method is helpful to magnify the spatial information of micro-expression, while remaining the temporal integrity of the clip such that the number of frames stays unchanged. This method was proved to be useful to improve the recognition performance in [47] [51]. The hyperparameter α of this method is set to 15. An augmentation we also conduct here is *cropping*, where for each video clip, each frame will be first resized to $144 \times 144$ and then cropped at random positions to make new images with size of $128 \times 128$. Data left for testing is directly resized to $128 \times 128$ without any extra processing. In short, the size of image used for all methods is $128 \times 128$.

During training, the number of video clips of each class will be equalized by resampling them according to the class comprising largest number of clips. Here we need to note that the length of each video clip remains unchanged by default, while further normalization on the video input length is used according to the original setting of each compared method.

TABLE I  DETAILS OF THE FOUR BENCHMARK DATASETS

| Feature | CASME I | CASME II | SMIC | SAMM |
|---|---|---|---|---|
| Clips number | 189 | 255 | 164 | 159 |
| Camera speed | 60fps | 200fps | 100fps | 200fps |
| Frame number | 9~86 | 24~141 | 11~58 | 30~101 |
| Face resolution | 167×137 | 257×225 | 260×213 | 374×343 |
| Categories | Happiness, Surprise, Tense, Disgust, Sadness, Fear, Repression, Contempt | Happiness, Surprise, Disgust, Sadness, Fear, Repression, others | Positive, Negative, Surprise | Happiness, Surprise, Anger, Disgust, Sadness, Fear, Contempt, |

---

[*]Code available at https://github.com/Trunpm/AKMNet-Micro-Expression

TABLE II ARCHITECTURE OF THE ADOPTED BACKBONE OF RESNET-18

| Layer | Kernel Parameter | Output Size |
|---|---|---|
| Conv1 | $5 \times 5, 64$ | $64 \times 64 \times 64$ |
| Conv2 x | $\begin{bmatrix} 3 \times 3, 64 \\ 3 \times 3, 64 \end{bmatrix} \times 2$ | $64 \times 32 \times 32$ |
| Conv3 x | $\begin{bmatrix} 3 \times 3, 128 \\ 3 \times 3, 128 \end{bmatrix} \times 2$ | $128 \times 16 \times 16$ |
| Conv4 x | $\begin{bmatrix} 3 \times 3, 256 \\ 3 \times 3, 256 \end{bmatrix} \times 2$ | $256 \times 8 \times 8$ |
| Conv5 x | $\begin{bmatrix} 3 \times 3, 512 \\ 3 \times 3, 512 \end{bmatrix} \times 2$ | $512 \times 4 \times 4$ |

*B. Experiment Set-Up*

ResNet-18 [31] is adopted for spatial feature extraction in the proposed AKMNet. The detailed architecture of the adopted ResNet-18 is shown in Table II. Given a video clip input $\mathbf{V} = \{I_1, I_2, ..., I_T\}$ with length $T$, the output is $F \in \mathbb{R}^{T \times 512 \times 4 \times 4}$. In order to alleviate the overfitting risk and aid the convergence of the training with micro-expression data, an initialization is conducted for the ResNet-18 using ImageNet dataset [37], which is further pre-trained with a macro-expression dataset named AffectNet [39]. For the four categories of positive, negative, supervise and others, we each randomly sample 1000 pictures from this dataset to conduct the pre-training.

For the adaptive key-frame mining module, a key-frame feature subset $F^{Key} = \{P_{k_1}, P_{k_2}, ..., P_{k_N}\}$ is generated, with $1 \leq k_1 < k_N \leq T$. The number of key frames $N$ is decided through network learning, and is not dependent on the input video length of $T$. Due to individual differences, even for the same micro-expression type, video clips with the same length may finally harvest key-frame subsets of different lengths. For Equation 12, $\lambda_1$ and $\lambda_2$ are set to 0.1 and 1, respectively.

For the spatio-temporal learning part, at each spatial position across all the frames, a two-layer bi-GRU network is used. Namely, the input for each bi-GRU network is a feature map with size of $N \times 512$, while the respective output dimension is $C' = 64$. The output of all the timestep of such network is averaged to be the output of each spatial position. The output of all the bi-GRU networks is concatenated together to form the spatio-temporal feature matrix $G \in \mathbb{R}^{64 \times 4 \times 4}$. This feature matrix is flattened into a vector and used as an input to the final fully-connected softmax layer for classification. A drop-out layer with probability of 0.5 is added before the softmax layer.

For the training of AKMNet, we set the mini-batch size to 8 and initial learning rate to 1e-3. The learning rate is adapted using a scheme named cosine annealing LR [29], with a minimum set to 1e-8. The number of maximum epochs is set to 40. For the back propagation of networks, the stochastic gradient descent (SGD) [30] is used, with momentum set to 0.9 and weight decay set to 5e-4. The leave-one-subject-out cross validation (LOSO) is used for all the methods.

For LBP-TOP, LBP-SIP, STCLQP, HIGO, and FHOFO, the number of frames of all the video clips remains unchanged. Specially, for LBP-TOP, LBP-SIP, and STCLQP, the radius of the X, Y, and temporal axis are set to 3 with the neighbor size set to 8. For LBP-TOP, the uniform LBP implementation is used to extract the feature. For STCLQP, the level of orientation estimation N and quantization level T are set to 16 and 4, respectively. For HIGO, the number of bins is set to 8. For FHOFO, we extract the optical-flow information based on [40], while the number of fine-grained bins is set to 36, the coarse-grained bins is set to 8 and the variance value is set to 10. For MDMO, the number of frames of each video clip composed CASME I is normalized to 64, and the video clip in other three datasets is normalized to 128, while for each frame the facial area is equally divided into 36 regions based on the original study. Similarly, for this method, the optical-flow feature is extracted using the implementation proposed in [40]. For the above methods, the linear SVM is used as classifier in the end.

*C. Comparison Experiment*

Results for the comparison between proposed AKMNet and other state-of-the-art methods are given in Table III. The first part of comparison methods were based on feature engineering and shallow classifiers. The five methods in the middle of the table used different deep learning architectures on the apex frame or the frames adjacent to the apex frame, or the onset frame. For CASME I/II and SAMM, annotations of the apex and onset frames were manually conducted, which are absent in SMIC. For [47] and [50], locations of the apex frames in SMIC are decided given the methods used in their original studies, respectively. In general, the best performance is achieved by the proposed AKMNet (accuracies of 0.7566, 0.6706, 0.7256, and 0.7170 on the four datasets, respectively), operated on video clips without using prior knowledge about the temporal property (e.g., location of the apex frame).

It is observed on CASME I and SAMM that, the deep learning methods based on the apex frame alone lead to generally better performance than traditional approaches using full video clips. However, such advantage is not seen on CASME II, where FHOFO achieved a better result of 0.6471 than Macro2Micro and MicroAttention methods. One explanation could be that, the informative temporal patterns of micro-expression were better captured in CASME II, which could be ignored by solely relying on the apex frame. A marginally better result is achieved with ATNet (accuracy of 0.6543) on SAMM in comparison with methods solely dependent on the apex frame, for which frames adjacent to the apex frame were added during modeling. This adds to the importance of temporal information for micro-expression recognition. Accordingly, the best result achieved by AKMNet can be owe to its adaptive key-frame mining module, which can search for an optimal temporal subset for the recognition. The confusion matrices of this method are shown in Figure 4. We make efforts to further justify this in the next subsection.

TABLE III RECOGNITION RESULTS (%) ON CASME I/II, SMIC, AND SAMM DATASETS IN THE COMPARISON EXPERIMENT

| Methods | CASME I | CASME II | SMIC | SAMM |
|---|---|---|---|---|
| LBP-TOP [8] | 0.6618 | 0.3843 | 0.3598 | 0.3899 |
| LBP-SIP [9] | 0.6026 | 0.4784 | 0.3842 | 0.5220 |
| STCLQP [34] | 0.6349 | 0.5922 | 0.5366 | 0.5283 |
| HIGO [10] | 0.5781 | 0.5137 | 0.3720 | 0.4465 |
| FHOFO [35] | 0.6720 | 0.6471 | 0.5366 | 0.6038 |
| MDMO [12] | 0.6825 | 0.6314 | 0.5793 | 0.6164 |
| Macro2Micro [19] (apex[*]) | 0.6772 | 0.6078 | - | 0.6436 |
| MicroAttention [20] (apex) | 0.6825 | 0.6431 | - | 0.6489 |
| ATNet [27] (apex+adjacent) | 0.6720 | 0.6039 | - | 0.6543 |
| STSTNet [50] (onset+apex) | 0.6349 | 0.5529 | 0.5488 | 0.6289 |
| STRCN-G [47] (onset+apex) | 0.7090 | 0.6039 | 0.6280 | 0.6478 |
| **AKMNet** | **0.7566** | **0.6706** | **0.7256** | **0.7170** |

[*]The input for the model, with apex= the apex frame, adjacent=frames adjacent to the apex frame, onset=the onset frame.

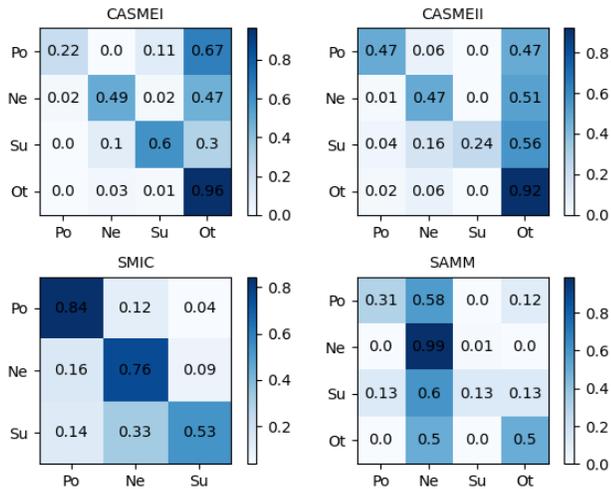

Fig. 4. Confusion matrices of AKMNet on the four datasets. Po=positive, Ne=negative, Su=surprise, Ot=others.

### D. Justification of the Adaptive Key-Frame Mining Module

In order to justify the effectiveness and robustness of the proposed AKMNet in terms of its adaptive key-frame mining design, we conduct a justification experiment here. We use **AKMNet$^{va}$** to denote a dedicated variant of AKMNet where the only difference is the absence of the adaptive key-frame mining module. That is, spatial feature sequence $F$ generated by ResNet-18 is processed by the bi-GRU networks directly. In detail, variants we designed here includes:

- **AKMNet$^{va}$-all**, where all the frames in an input video clip are used.
- **AKMNet$^{va}$-norm$\beta$**, where an input video clip is first normalized with TIM [8] to be a new clip with frames number $\beta \in \{16,32,64,128\}$.
- **AKMNet$^{va}$-random**, where for each input video clip, given the number of key frames $N$ learned by the default AKMNet, respectively, the same number of frames is randomly sampled to be the input of this model. The result reported here is the average performance after five repeated random samplings.

Results are reported in Table IV. With the increase of $\beta$ from 16 to 128, the performance of AKMNet$^{va}$-norm$\beta$ is improved on CASME II and SMIC datasets that both have a framerate ≥ 100 fps. However, the pattern is less obvious for the results on SAMM dataset, which also has a high framerate of 200 fps though. A marginal decrease of performance is seen on CASME I, when the number of frames is normalized as 128. This is even higher than the original framerate of the CASME I dataset. These results suggest that the temporal information of micro-expression could be disturbed by the improper use of temporal interpolation, e.g. not considering the framerate of the videos. On the other hand, given a proper $\beta$, AKMNet$^{va}$-norm$\beta$ leads to better performance than the one using all the frames of a video clip as input (AKMNet$^{va}$-all) in all the datasets. Such reveals the temporal redundancy existed in the video clip. A discussion about generating fixed number of frames using TIM is seen in several works [10] [12] [13] [18]. However, given the individual differences in the presentation of micro-expression, the length of underlying optimal temporal subset of micro-expression should be varying.

To the best of our knowledge, this study is the first to leverage the temporal characteristic of micro-expression under different temporal scales. In this experiment, the best performance is achieved by the proposed AKMNet that adaptively learns to find the optimal subset of frames in the input video clip for micro-expression recognition.

### E. Ablation Experiment

The function of the adaptive key-frame mining module is fulfilled through three steps, namely local self-attention learning, global correlation learning, and sparse selection. It can be noticed that all these steps have some overlap on each other, while it remains unknown if our proposal of combining the three is the best approach for the key-frame mining purpose. Here, we run an ablation study on the three steps by creating three meaningful variants of the AKMNet:

- **AKMNet-s12**, where the sparse selection step is absent.
- **AKMNet-s13**, where the global correlation learning step is absent.
- **AKMNet-s23**, where the local self-attention learning step is absent.

The proposed method used in other subsections is now referred to as **AKMNet-s123**, as to help the discrimination among them. Results are reported in Table V. We can see that, by removing one component from the proposed key-frame mining module, all the three variants lead to reduced performance than our proposed method. Such result justified our use of a combination of the three steps. In all the datasets, variants equipped with step three of sparse selection (*AKMNet-s13* and *AKMNet-s23*) have competitive if not better results than the one without it (*AKMNet-s12*). This result may once again validate that the key frames of micro-expression are sparsely spread in a video clip, while a sparse selection step could help the network to get rid of redundant frames instead of reserving them with lower weights.

TABLE IV RECOGNITION RESULTS (%) ON CASME I/II, SMIC, AND SAMM DATASETS IN THE JUSTIFICATION EXPERIMENT

| Methods | CASME I | CASME II | SMIC | SAMM |
| --- | --- | --- | --- | --- |
| AKMNet$^{va}$-all | 0.6720 | 0.6118 | 0.5793 | 0.6415 |
| AKMNet$^{va}$-random | 0.6138 | 0.6118 | 0.5427 | 0.6289 |
| AKMNet$^{va}$-norm16 | 0.6667 | 0.6314 | 0.5976 | 0.6604 |
| AKMNet$^{va}$-norm32 | 0.6825 | 0.6392 | 0.6434 | 0.6478 |
| AKMNet$^{va}$-norm64 | 0.7090 | 0.6392 | 0.6463 | 0.6164 |
| AKMNet$^{va}$-norm128 | 0.6984 | 0.6431 | 0.6646 | 0.6792 |
| **AKMNet** | **0.7566** | **0.6706** | **0.7256** | **0.7170** |

TABLE V RECOGNITION RESULTS (%) ON CASME I/II, SMIC, AND SAMM DATASETS IN THE ABLATION EXPERIMENT

| Methods | CASME I | CASME II | SMIC | SAMM |
| --- | --- | --- | --- | --- |
| AKMNet-s12 | 0.6984 | 0.6392 | 0.6463 | 0.6667 |
| AKMNet-s13 | 0.7354 | 0.6431 | 0.6463 | 0.6604 |
| AKMNet-s23 | 0.7249 | 0.6549 | 0.6707 | 0.6918 |
| **AKMNet-s123** | **0.7566** | **0.6706** | **0.7256** | **0.7170** |

### F. Is the Annotated Apex Frame the 'Most Informative'?

So far, we have shown the advantage of the proposed AKMNet in comparison with methods that directly leverage the expert-annotated apex frame or onset frame as input. There lacks another important study to understand the relationship between the key frames learned by the AKMNet and the respective apex frame annotated by human experts. It was supported by a series of studies [19] [20] [21] [22] [27] that, for a micro-expression video clip, the apex frame shall contribute most to the recognition performance. Given such, the key frames learned by the AKMNet for each video clip may also comprise the apex frame and some other relevant frames.

In this subsection, we look into the key frames learned by AKMNet from each micro-expression video clip to understand such interplay. The CASME I/II and SAMM datasets are used, given the annotations of onset, offset, and apex frames are available. Firstly, for each dataset, we calculate the ratio of video clips where i) the apex frame is included in the sequence of key frames learned by the AKMNet (denoted as *Ratio*); ii) the apex frame is included and also assigned with the highest weight ($\beta_t$) among the key frames (denoted as $Ratio^W$). Results are reported in Table VI. We can see that, for most video clips (>50%) of the three datasets, the apex frame annotated by human experts is also one of the key frames learned by AKMNet. However, it is interesting to see that there are only ~1% video clips in CASME II and SAMM datasets where the apex frame is assigned with the highest weight, and such ratio is higher (~8%) in CASME I dataset. In other words, to a large extent, the proposed data-driven AKMNet does not find the apex frame to be the most informative for micro-expression recognition. This is possibly due to the error that human experts had committed during their apex-frame annotations, resulting in a temporal distance (a number of frames) between the apex frame and underlying most-informative key frame. Additionally, human experts may perceive the micro-expression in a slightly different way from a data-driven model. The latter only aims to increase its recognition performance by searching for informative frames.

Thereon, we empirically verify if the key frame assigned with the highest weight (denoted as *max-key frame*) learned by AKMNet is more informative than the apex frame annotated by human experts, from a recognition perspective. Another experiment with vanilla models of ResNet-18 [31] and VGG-11 [52] is conducted. The former is the backbone of Macro2Micro model [19] and micro-expression recognizer [43], while the latter is used in DTSCNN [18] and STSTNet [50]. Under the same experimental setting, we run each model with the apex frame and the max-key frame (excluding the situation when apex frames are given the highest weight) as input separately on the three datasets. Results are reported in Table VII. As we can see, better performances are achieved using the max-key frame produced by AKMNet for all the three datasets, in comparison with the expert-annotated apex frame. Such results empirically verify the existence of a very informative frame within each video clip, which however could get ignored by human experts during the annotation. Here, we need to clarify that such experiment is not challenging the idea of using the apex frame for micro-expression recognition, but suggesting the expert annotation of such frame may not be that accurate, from a recognition perspective. In Figure 5, we show the distribution of the distance between the apex and max-key frames in each video clip per datasets. Given the larger framerates of CASME II and SAMM datasets in comparison with CASME I (200 vs. 60 fps), distances computed for the two are normalized by dividing with a factor of 3.33. We can see that most apex frames are located within the distance of less than 10 frames (1/6 second) from the respective max-key frames. This may imply again the marginal error that human experts could commit during the annotation.

Recently, there are also some studies [44] [45] that particularly focused on micro-expression spotting, where the aim is to spot the onset, offset, and apex frames in comparison with expert annotations. It should be noted that, unlike micro-expression spotting, the key frame mining module proposed in this paper is unconstrained to those frames. In other words, the number of key frames is not predefined, and it is the recognition performance rather than the inclusion of those frames that being optimized during training.

TABLE VI  RATIOS (%) OF VIDEO CLIPS IN CASME I/II AND SAMM DATASETS WHERE THE APEX FRAME IS LEARNED AS THE KEY FRAME, AND ASSIGNED WITH THE HIGHEST WEIGHT

| Dataset | Ratio | $Ratio^W$ |
|---|---|---|
| CASME I | 0.6643 | 0.084 |
| CASME II | 0.5439 | 0.012 |
| SAMM | 0.5088 | 0.018 |

TABLE VII  RECOGNITION RESULTS (%) OF MODELS WITH INPUT OF APEX FRAMES AND MAX-KEY FRAMES, RESPECTIVELY

| Backbone | Input | CASME I | CASME II | SAMM |
|---|---|---|---|---|
| ResNet-18 | Apex frame | 0.6772 | 0.6078 | 0.6436 |
|  | Max-key frame | **0.6825** | **0.6392** | **0.6486** |
| VGG-11 | Apex frame | 0.6667 | 0.6235 | 0.6277 |
|  | May-key frame | **0.6931** | **0.6353** | **0.6649** |

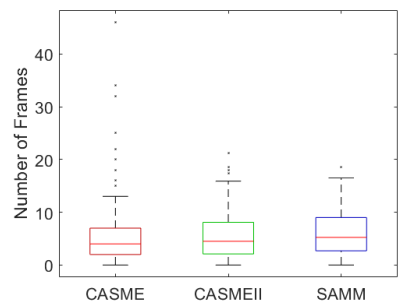

Fig. 5. Boxplots of the distribution of the distance between max-key and apex frames in each video clip of CASMEI/II and SAMM datasets.

### G. Generalization Experiment

Experiments conducted above have shown the effectiveness of the proposed approach on micro-expression recognition with video clips. The design of AKMNet is not only limited to the recognition of micro-expression, but also applicable to the more generic scenario where the events-of-interest are sparsely located in a data sequence. Here, we adopt a facial macro-expression benchmark dataset CK+ [28] to test the applicability of AKMNet beyond the micro-expression recognition domain. In the CK+ dataset, 327 video clips captured from 118 participants were labelled with 7 macro-expression categories, namely happiness, anger, sadness, fear, disgust, contempt, and surprise. In our experiment, we run a hold-out validation where data from 80% of the participants are used for training with the rest 20% participants left out for testing. Five variants of the AKMNet created in the justification experiment are used here to aid the comparison, namely AKMNet$^{va}$-all and AKMNet$^{va}$-norm$\beta$, $\beta\epsilon\{16,32,64,128\}$. In addition, another variant named AKMNet$^{va}$-Last10 is created, where the input to the network is the last 10 frames of each video clip. Such was adopted in the original study [28], because for each video clip the intensity of the macro-expression increases over time. Results are summarized in Table VIII. The best performance is still achieved by the proposed AKMNet, in comparison with the methods taking in all the input frames, or applying the temporal interpolation at different scales, or only using the last 10 frames. This result shows the generalization ability of the proposed approach on a relevant scenario, even if some intrinsic characteristics of the target are not fully leveraged during the modeling e.g., the larger size of available data and higher spatial intensify of the facial macro-expression. We believe higher recognition performances can be achieved by modifying the AKMNet according to these factors. The method taking in all the input frames (AKMNet$^{va}$-all) has better performance than the ones using temporal interpolations (AKMNet$^{va}$-norm$\beta$, $\beta\epsilon\{16,64,128\}$). One reason for such result could be that the higher facial movement intensify of macro-expression left more continuous information across all the frames, which could be harmed by the use of temporal interpolation. On the other hand, the proposed method achieves a better performance than AKMNet$^{va}$-Last10, suggesting the existence of other informative frames for some video clips beyond the last 10 frames. It also shows the advantage of using an adaptive key-frame mining strategy in comparison with a manually predefined strategy that driven by the data nature, e.g., the increasing macro-expression intensity in this case.

## V. Conclusions

The recognition of micro-expression has reached a new era where researchers started to leverage the intrinsic characteristics of such facial expression to guide the modeling. Specifically, the subtle intensity and short duration of micro-expression have been harnessed during the method development. In this paper, we have focused on the transient presentation of micro-expression, which is usually sparsely located in a video clip. An end-to-end network architecture named AKMNet featuring an adaptive key-frame mining module has been proposed. The module was designed to learn an optimal temporal subset from the spatial feature sequence of an input video clip. Our evaluation with four benchmark datasets, namely CASMEI/II, SMIC, and SAMM has showed noticeable improvements achieved by the AKMNet over state-ot-the-art methods. A further justification study has verified the effectiveness of the proposed adaptive key-frame mining module, in comparison with the variants using all frames, or different temporal interpolations, or random samplings of the video clip. In the ablation study, we validated the advantage of the combination of the three steps in the key-frame mining module, which produced the best performance than other implementations. In the study comparing expert-annotated apex frame and model-learned max-key frame, we empirically showed the advantage of using the latter one as the single-frame input. This experiment also revealed the possible error that could be committed by human experts during data annotation, and also the perception difference between the human expert and a data-driven model (e.g., AKMNet in our case). Finally, a comparable generalization ability of the AKMNet on a relevant macro-expression benchmark dataset was also witnessed. One limitation of this work is the dependence on ResNet-18 for spatial feature extraction, which is convenient to use in an end-to-end network but not able to cover facial features related to action units and facial landmarks. We leave the exploration of integrating informative facial features into the network to future work.
...ACKNOWLEDGMENTThis work is funded by the National Natural Science Foundation of China (Grant No. 61802361 and Grant No. 61806185). Chongyang Wang is supported by the UCL Overseas Research Scholarship (ORS) and Graduate Research Scholarship (GRS).
## Acknowledgment

This work is funded by the National Natural Science Foundation of China (Grant No. 61802361 and Grant No. 61806185). Chongyang Wang is supported by the UCL Overseas Research Scholarship (ORS) and Graduate Research Scholarship (GRS).

TABLE VIII RECOGNITION RESULTS (%) ON CK+ DATASET IN THE GENERALIZATION EXPERIMENT

| Methods | Accuracy |
|---|---|
| AKMNet$^{va}$-all | 0.8950 |
| AKMNet$^{va}$-norm16 | 0.8895 |
| AKMNet$^{va}$-norm32 | 0.8979 |
| AKMNet$^{va}$-norm64 | 0.8347 |
| AKMNet$^{va}$-norm128 | 0.8681 |
| AKMNet$^{va}$-Last10 | 0.9165 |
| **AKMNet** | **0.9336** |

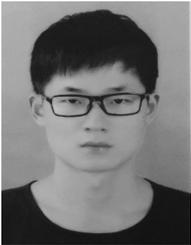

**Min Peng** is a Ph.D. student at Chongqing School, University of Chinese Academy of Sciences, China. He is also a research fellow at Intelligent Security Center, Chongqing Institute of Green and Intelligent Technology, Chinese Academy of Sciences, China. In 2017, he received the M.S. degree in affective computing from Southwest University, China. His research interests include machine learning, computer vision and affective computing.

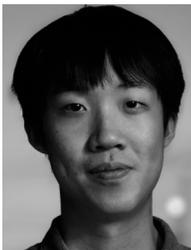

**Chongyang Wang** is a Ph.D. candidate at UCL interaction centre, University College London, United Kingdom, under the supervision of Prof. Nadia Bianchi-Berthouze and Dr. Nicholas D. Lane. In 2017, he received the B.E. degree in electronic and information engineering from Southwest University, China. His current research topics include machine learning and developing ubiquitous body sensing technology to support physical rehabilitation of people with chronic pain.

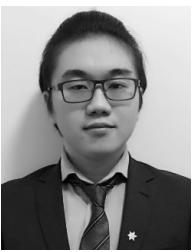

**Yuan Gao** is an assistant research scientist at Shenzhen Institute of Artificial Intelligence and Robotics for Society. He received his Ph.D. from Department of Information Technology, Uppsala University, Sweden under the supervision of Prof. Ginevra Castellano, and Prof. Danica Kragic Jensfelt. In 2016, he received his M.S. degree in Algorithms and Machine Learning from University of Helsinki, Finland. His current research topics include deep reinforcement learning for robotics and deep learning-based approaches for perception and control robotics.

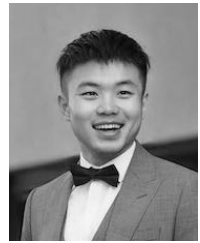

**Tao Bi** is a Ph.D. candidate under the supervision of Prof. Nadia Bianchi-Berthouze at UCL interaction centre, University College London, United Kingdom, where he also obtained his M.S. Degree in HCI in 2017. His research interests include Human-Computer Interaction and Affective Computing.

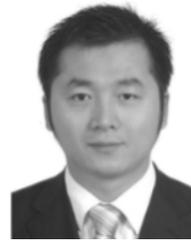

**Tong Chen** received the B.E. degree in communication engineering from The Second Artillery Command Institute, China, in 2002, the M.S. degree in digital sound and vision processing from University of Wales, United Kingdom, in 2006, and the Ph.D. degree from Cranfield University, United Kingdom, in 2013. He is currently a professor in Southwest University, China. His research interests include image/signal processing, machine learning, and affective computing.

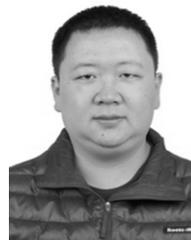

**Yu Shi** is the director of Center on Research of Intelligent Security Technology, CIGIT, Chinese Academy of Sciences, China. He leads a team for core technology and industrial application research in Computer Vision and Pattern Recognition area.

He has published more than 20 patents, and has obtained 4 patent licenses. He is the West Light A Class awarded by Chinese academy of sciences.

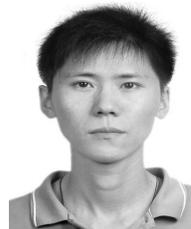

**Xiang-Dong Zhou** is an associate professor at CIGIT, Chinese Academy of Sciences, China. He received the B.S. degree in Applied Mathematics and the M.S. degree in Management Science and Engineering both from National University of Defense Technology, China, the Ph.D. degree in pattern recognition and artificial intelligence from the Institute of Automation, Chinese Academy of Sciences, China, in 1998, 2003 and 2009, respectively.

He was a postdoctoral fellow at Tokyo University of Agriculture and Technology, Japan from March 2009 to March 2011. His research interests include handwriting recognition and ink document analysis.